\documentclass{bmvc2k}
\pdfoutput=1
\usepackage{booktabs}       
\usepackage{amsfonts}       
\usepackage{nicefrac}       
\usepackage{microtype}      
\usepackage{amsmath,amssymb,amsthm}
\usepackage{mathtools}
\usepackage{etoolbox}
\usepackage{enumitem}
\usepackage{tabularx}
\usepackage[group-separator={,}]{siunitx}
\usepackage[ruled,vlined]{algorithm2e}
\usepackage{subcaption}
\usepackage{graphicx}
\usepackage{floatrow}
\usepackage{rotating}

\newfloatcommand{capbtabbox}{table}[][\FBwidth]
\newcommand{\projectPage}[1]{\url{https://github.com/ashafaei/OD-test}}

\title{A Less Biased Evaluation of Out-of-distribution Sample Detectors}

\addauthor{Alireza Shafaei}{http://www.shafaei.ca}{1}
\addauthor{Mark Schmidt}{http://cs.ubc.ca/~schmidtm}{1}
\addauthor{James J. Little}{http://cs.ubc.ca/~little}{1}

\addinstitution{
 Computer Science Department\\
 University of British Columbia\\
 Vancouver, Canada
}

\runninghead{Shafaei \etal{}}{A Less Biased Evaluation of OOD Sample Detectors}
\def\eg{\emph{e.g}\bmvaOneDot}

\def\etal{\emph{et al}\bmvaOneDot}

\graphicspath{{figures/}}

\newcommand{\D}[1]{\ensuremath{\mathcal{D}_{\mathrm{#1}}}}
\newcommand{\Ds}[2]{\ensuremath{\mathcal{D}_{\mathrm{#1}}^\mathrm{#2}}}
\begin{document}
\newgeometry{twoside,headsep=3mm,papersize={410pt,620pt},inner=14mm,outer=6mm,top=3mm,includehead,bottom=5mm,heightrounded}

\maketitle

\begin{abstract}
In the real world, a learning system could receive an input that is unlike anything it has seen during training.
Unfortunately, out-of-distribution samples can lead to unpredictable behaviour.
We need to know whether any given input belongs to the population distribution of the training/evaluation data to prevent unpredictable behaviour in deployed systems.
A recent surge of interest in this problem has led to the development of sophisticated techniques in the deep learning literature.
However, due to the absence of a standard problem definition or an exhaustive evaluation, it is not evident if we can rely on these methods.
What makes this problem different from a typical supervised learning setting is that the distribution of outliers used in training may not be the same as the distribution of outliers encountered in the application.
Classical approaches that learn inliers vs. outliers with only two datasets can yield optimistic results.
We introduce OD-test, a three-dataset evaluation scheme as a more reliable strategy to assess progress on this problem.
We present an exhaustive evaluation of a broad set of methods from related areas on image classification tasks.
Contrary to the existing results, we show that for realistic applications of high-dimensional images the previous techniques have low accuracy and are not reliable in practice.
\end{abstract}

\section{Introduction}
\label{sec:intro}
\begin{figure*}
    \centering
    \includegraphics[width=0.95\textwidth]{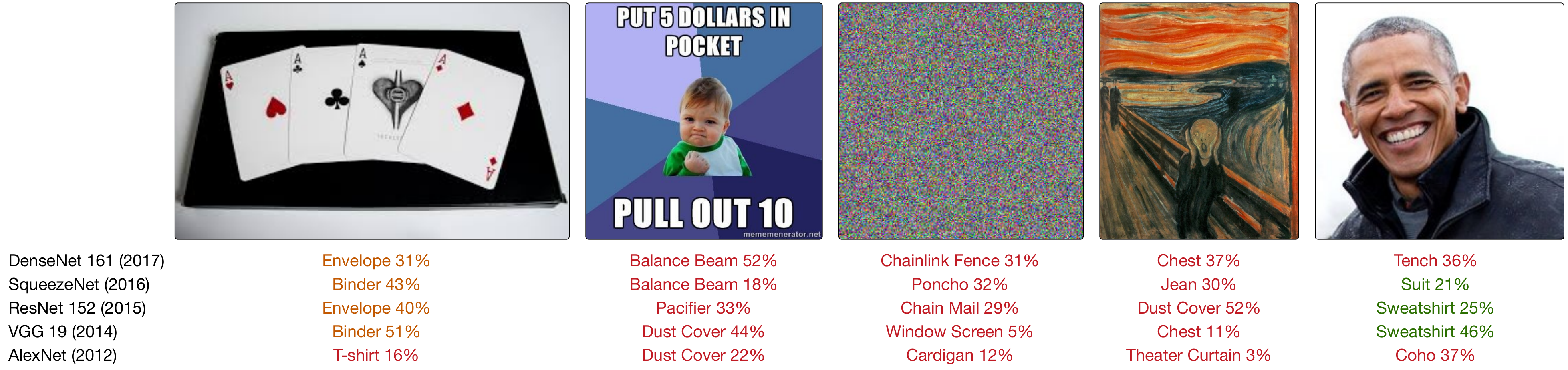}    
    \caption{
        The predictions of several popular networks~\citep{Huang2017,Iandola2016,He2016,Simonyan2014,Krizhevsky2012} that are trained on ImageNet on unseen data. The red predictions are entirely wrong, the green predictions are justifiable, and the orange predictions are less justifiable. The middle image is Gaussian noise. We show that thresholding the output probability is not a reliable defence.
        }
    \label{fig:arbit_imagenet}
\end{figure*}

If we present a natural image of an unknown class to the currently popular deep neural network that is trained to discriminate ImageNet~\citep{Russakovsky2015} classes, we will get a prediction with a high (softmax) probability of an arbitrary class (see Fig.~\ref{fig:arbit_imagenet}). With English speaking phone assistants, if we talk in another language, it will generate an English sentence that most often is not even remotely similar to what we have said. The silent failure of these systems is due to an implicit assumption: the input to the ImageNet classifier \textit{will be} from the same ImageNet distribution, and the user \textit{will be} speaking in English. However, in practice, any automation pipeline that involves a deep neural network will have a critical challenge:

\vspace{-0.1cm}
\begin{center}
    \emph{Can we trust the output of a neural network for a particular input?}
\end{center}
\vspace{-0.1cm}
\newgeometry{twoside,headsep=3mm,papersize={410pt,620pt},inner=9mm,outer=6mm,top=3mm,includehead,bottom=5mm,heightrounded}

One solution is to add a \texttt{None} class to the models to account for the absence of other classes. The first challenge is defining the \texttt{None} class. Would \texttt{None} mean all other image vectors or only vectors of natural images? Another non-trivial challenge is capturing the diversity of \texttt{None} class with a finite sample set to use for training. The third and the most prohibitive problem is that we have to significantly increase the complexity of our models to capture the diversity of the \texttt{None} class. Despite these challenges, we might be able to achieve reasonable results on low dimensional problems~\cite{hendrycks2018deep}, but as the input dimension grows, the problem becomes more severe. Our argument rests on the assumption that learning a reliable \texttt{None} class is not a viable strategy for practical applications.

In supervised learning we typically assume the samples are independently and identically distributed (IID) -- we expect the train and test examples to be drawn from a fixed population distribution. However, this condition cannot be easily enforced in deployed applications.
When the IID assumption is not satisfied, the empirical error can no longer predict the performance. Without any assumptions, the outputs on out-of-distribution (OOD) samples can be arbitrarily bad. The OOD samples violate the \textit{identically distributed} (ID) assumption. Thus, as long as we rely on the empirical error alone to train and evaluate deep neural networks, the first condition for making a reliable prediction (with bounded error) on any input would be whether the ID assumption is satisfied. When we \textit{a priori} expect to change the underlying distribution, the traditional applicable frameworks are transfer learning, multitask learning, and zero-shot learning. In our setting, we only wish to \textit{detect} out-of-distribution samples.

In real life deployment of products that use complex machinery such as deep neural networks (DNNs), we would have very little control over the input. In the absence of extrapolation guarantees, when the IID assumption is violated, the behaviour of the pipeline may be unpredictable. A reliable pipeline would first determine whether it can process a given sample, then it would use the prediction of the target neural network.
Successful detection of such violations could also be used in active learning, unsupervised learning, learning with noisy data, or simply be a condition to invoking transfer learning strategies.


There has been a recent surge of interest in specifically OOD sample detection with deep neural networks~\citep{Hendrycks2017,Liang2018,Schlegl2017,Bendale2016,Lakshminarayanan2017,hendrycks2018deep}. However, the commonly applied evaluation mechanisms are susceptible to overly optimistic results and cannot provide a conclusive evidence of reliability (we demonstrate this).
There is an imminent need to define a standard problem and properly define a benchmark that is both realistic and principled to compare the previous and future approaches reliably.

\vspace{0.4cm}
\noindent Our contributions are as follows:
\vspace{-0.1cm}
\begin{enumerate}
    \setlength\itemsep{-0.4em}
    \item We introduce OD-test: a more realistic formulation of the OOD detection task.
    \item We present a new benchmark to evaluate the existing techniques for OOD detection.
    \item We provide analysis of several previously proposed methods under OD-test.
    \item We release a PyTorch~\citep{Paszke2017} package to replicate all the results (\projectPage{}).
\end{enumerate}
\vspace{-0.1cm}
We demonstrate that the performance of current techniques quickly approaches the random prediction baseline as we make a transition to realistic high-dimensional images.

\section{Related Work}
\label{sec:rel_work}
\label{sec:related_work_all}
The violation of the ID assumption is not the only way to wreak havoc on deep learning pipelines.
Adversarial example~\citep{Szegedy2014} attacks are crafted signals in otherwise innocent-looking images that fool the neural networks into misclassification.
While the OOD detection is a model-independent problem, adversarial images exploit inductive bias in the model families.
We limit our attention to the OOD detection problem.

\vspace{-0.45cm}
\paragraph{The Uncertainty View} An uncertainty measure could be directly applied to reject OOD samples as we would expect the uncertainty to be high on such inputs. The \texttt{MC-Dropout}~\citep{Gal2016} approach is a feasible uncertainty estimation method for a variety of applications~\citep{Kendall2017,Gal2016,Gal2016b}. \citet{Lakshminarayanan2017} show an ensemble of five neural networks (\texttt{DeepEnsemble}) trained with an adversarial-sample-augmented loss is sufficient to provide a measure of predictive uncertainty. We evaluate \texttt{DeepEnsemble} and \texttt{MC-Dropout}.

\vspace{-0.45cm}
\paragraph{The Abstention View} If we abstain from prediction at the cost of a penalty, we end up with the abstention view. A reject function makes the abstention choice and it could be a threshold on magnitude of the prediction~\citep{Bartlett2008} or chosen from a reject-hypothesis set~\citep{Cortes2016,Cortes2017}. The reject function is chosen simultaneously with the predictive function through an abstention-augmented loss.
The reliability of such reject function is contingent on evaluation on a fixed distribution. If we encounter an OOD sample, we do not know for sure if it would be rejected. Our formulation of the OOD detection task is similar to the abstention view with key differences that we will discuss in Section~\ref{sec:formulation}. We show that the prior work on OOD detection can be reduced to an abstract problem of choosing a reject function from a specific space.

\vspace{-0.45cm}
\paragraph{The Anomaly View} \textit{Density estimation}-based techniques assume that low measure samples are outliers. These approaches tend to work well mostly within low-dimensional or well-defined distributions. \texttt{PixelCNN++}~\citep{Salimans2017} is an auto-regressive model with a tractable likelihood that could be used within a density estimation scheme. Note that density estimation is \textit{not equivalent} to the binary OOD detection: a perfect density estimator can solve the outlier detection problem, but a perfect outlier detector does not necessarily have the information needed to solve the density estimation problem. \textit{Proximity}-based methods use a distance measure and the train data to flag anomalies. A simple strategy is based on the K-nearest neighbours of a given input -- we call this \texttt{K-NNSVM}. Clustering methods reject points that do not conform to any of the identified clusters. The one-class SVM~\citep{Scholkopf2000} with a radial basis function learns a conservative region that covers the train data. \citet{Goldstein2016} show that the proximity-based approaches are empirically the most effective outlier detectors over a range of datasets. \textit{Reconstruction}-based methods learn to reconstruct the train data, then try to reconstruct each given input. The samples that cannot be reconstructed well are then flagged as anomalies. We use an autoencoder with a reconstruction threshold to test this idea (\texttt{AEThreshold}).

\vspace{-0.45cm}
\paragraph{The Novelty View} Open-set recognition and novelty detection study the detection of anomalies at a \textit{semantic} level. These methods are typically concerned with recognition of unseen classes, \eg{}, new objects in the scene. This is a special case of OOD detection where the OOD samples explicitly differ by the semantic content. However, the notion of OOD is more granular: an unseen viewpoint of a specific object violates the ID assumption, but it does not necessarily constitute a novelty. The notion of novelty is often underspecified in practice and results are limited to particular assumptions and problem definitions. \citet{Bendale2016} present \texttt{OpenMax}, a replacement for the softmax layer that detects unknown classes through evaluation against a representative neural activation of each class.

\paragraph{Deep Learning Literature}
\label{sec:dl_literature}
\citet{Guo2017} observed that neural networks tend to be overconfident in predictions. They show that temperature scaling is an effective calibration strategy for neural networks.
\citet{Hendrycks2017} show that it is possible to detect OOD samples by thresholding the softmax probabilities.
More recently, \citet{Liang2018} presented \texttt{ODIN}, a method based on temperature rescaling and input perturbation to detect OOD samples.
Further extensions rely on statistics of hidden representations~\citep{lee2018simple,quintanilha2019detecting}, construct classifier ensembles with subsets of data~\citep{vyas2018out}, or perform regression on word embeddings~\citep{shalev2018out}.

There are also ideas that rely on GANs~\citep{Deecke2018a,Kliger2018,Schlegl2017,Lee2018}
to detect anomalies or novelty in the data.
To the best of our knowledge, training GANs~\citep{Goodfellow2014a} is an active area of research, and it is not apparent what design decisions would be appropriate to implement these ideas in practice.
We are therefore unable to evaluate these ideas fairly at this time.

All the previous studies primarily focus on low-dimensional MNIST~\citep{LeCun1998}, SVHN~\citep{Netzer2011}, and CIFAR~\citep{Krizhevsky2009} datasets. We evaluate several previously proposed solutions in controlled experiments on datasets with varying complexity. We show that, in such low-dimensional spaces, simple anomaly detection methods work as well, thus stressing that a more comprehensive evaluation is necessary for assessments of the current and future work.

\section{OD-test: A Less Biased Evaluation of Outlier Detectors}
\label{sec:formulation}

Let us define the source distribution \D{s} to be our input distribution. The objective is to decide whether a given sample belongs to \D{s}. We define a reject function $r:\mathcal{X}\rightarrow \{0,1\}$ that makes this binary decision. Note that this decision can be made independently from the ultimate prediction task. While in the abstention view we reject the samples that the predictive function is likely to mislabel, here we reject the samples that do not belong to the source distribution \D{s}, hence decoupling the reject function and the predictive function.

If the reject function flags an input, then the sample does not belong to the source distribution; thus, the output of the pipeline may not be reliable. On the other hand, if the function accepts an input, we can continue the pipeline with the ID assumption. This form of rejection is more general than the previous work. In addition to the previous methods, we can study new approaches that operate in the input space directly (\eg{} \texttt{K-NNSVM}, \texttt{AEThreshold}).

The $r$ function is a binary classifier; the classes are in-distribution vs. out-of-distribution.
To learn the classifier, the standard approach is to adopt the supervised learning assumptions and use an outlier dataset \D{v}.
In that scenario, what would happen if an outlier that we encounter is not represented by \D{v}?
We end up with the original problem again.
The traditional supervised outlier detection may overestimate our ability to detect outliers.
A high accuracy in this scenario may not yield an accurate model for practice in many settings where the outlier may not look like samples from \D{v}.
The actual OOD samples are beyond our direct reach and our models can easily overfit in distinguishing \D{s} from \D{v} (we verify this empirically, see Fig.~\ref{fig:eval_comp}). We present a less optimistic evaluation framework that prevents scoring high through overfitting.

\begin{figure}
\ffigbox[\textwidth]
{
    \begin{floatrow}
    \ffigbox[0.8\linewidth]{%
        \captionof{subfigure}{}\label{fig:alg_vis}
    }{
        \includegraphics[width=\columnwidth]{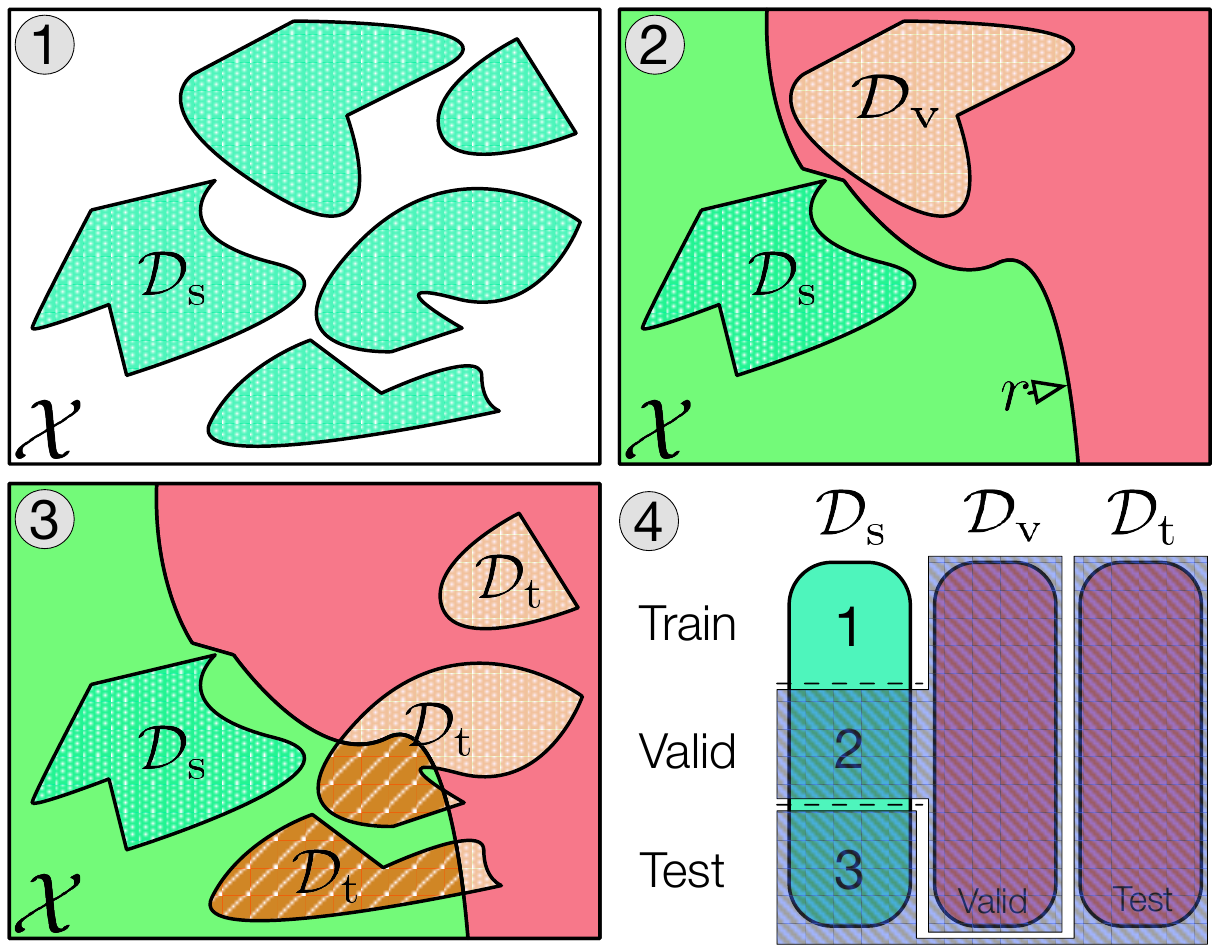}
    }
    \capbtabbox[1.1\linewidth]{%
        \resizebox{\columnwidth}{!}{%
        \begin{tabular}{@{}lccccrl@{}}
            \toprule
            & \multicolumn{3}{c}{\D{s} -- Source} &
              \multicolumn{1}{c}{Outliers} \\
            \cmidrule(r{4pt}){2-4} \cmidrule(l){5-5}
            & Train & Valid & Test & All & $\dim(\mathcal{X})$ & $|\mathcal{Y}|$\\
            \midrule
            MNIST              & (50\,k & 10\,k) & 10\,k & 70\,k & 784 & 10 \\
            FashionMNIST        & (50\,k & 10\,k) & 10\,k & 70\,k & 784 & 10 \\
            NotMNIST & & & & 18.6\,k & 784 & 10 \\
            CIFAR10       & (40\,k & 10\,k) & 10\,k & 60\,k & \num{3072} & 10 \\
            CIFAR100      & (40\,k & 10\,k) & 10\,k & 60\,k & \num{3072} & 100 \\
            TinyImagenet & 100\,k & 10\,k & 10\,k & 110\,k & \num{12288} & 200 \\ 
            STL10 & 5\,k & (4\,k & 4\,k) & 13\,k & \num{27648} & 10\\
            \bottomrule
          \end{tabular}
          }
    }{
        \captionof{subfigure}{}\label{tab:ds_summary}
    }
    \end{floatrow}
}{
\caption{The OD-test. a.1) A set of distributions visualized as shapes within $\mathcal{X}$. The source distribution \D{s} is identified in the image; everything else is an outlier. a.2) We pick one validation distribution \D{v} and learn a binary reject function $r$ that partitions the input space $\mathcal{X}$ based on \D{s} and \D{v} only. a.3) We evaluate $r$ on other distributions (as \D{t}) and measure the accuracy. a.4) The dataset splits for each step. b) A summary of the datasets. The datasets are split as indicated by the parentheses.}\label{fig:hotspot}
}
\end{figure}

Specifically, we introduce a third ``target'' distribution \D{t} to measure whether a method can actually detect outliers that are not only outside of \D{s} but that also might be outside of \D{v}. The idea is to treat the problem as a binary classification between three different datasets. Similar to the supervised outlier detection, we begin by learning a reject function to distinguish \D{s} from \D{v}. For evaluation, we use a third unseen distribution \D{t} instead of \D{v} (see Fig.~\ref{fig:alg_vis}). \D{t} represents OOD examples that were not encountered during training -- a more realistic evaluation setting for uncontrolled scenarios.
Ideally, \D{v} and \D{t} should have no similarity. In practice, we recommend choosing a collection of datasets with no label overlap.
In our experiments, we cycle through all choices of outliers (\D{v}, \D{t} from Tab.~\ref{tab:ds_summary}) and average the results. The pseudocode of the evaluation procedure is outlined in Alg.~\ref{alg:eval}.

\addtolength{\textfloatsep}{-0.15in}

\begin{algorithm}
\scriptsize
\DontPrintSemicolon
\LinesNumbered
\SetKwInOut{Input}{input}
\SetKwInOut{Output}{}
\SetKwInOut{OptInput}{input}
\Input{$\mathcal{D}_{\mathrm{s}}=(\mathcal{D}_{\mathrm{s}}^\mathrm{train},\mathcal{D}_{\mathrm{s}}^\mathrm{valid},\mathcal{D}_{\mathrm{s}}^\mathrm{test})$ the source dataset.}
\Input{$\mathrm{D}=\{\mathcal{D}_i\}$ outlier \textbf{set}.}
\Input{$\mathcal{M}:\mathcal{D}\rightarrow\mathcal{R}$ the method under evaluation.}
\Begin{
$A \longleftarrow \{\}$\;
\tcc{Generate a reject-hypothesis class $\mathcal{R}$ using $\mathcal{D}_{\mathrm{s}}^\mathrm{train}$.}
$\mathcal{R} \longleftarrow \mathcal{M}(\mathcal{D}_{\mathrm{s}}^\mathrm{train})$\;\label{alg:code:m_train}
\For{$\mathcal{D}_{\mathrm{v}}\in \mathrm{D}$}{
\tcc{Find the best binary classifier in $\mathcal{R}$.}
$r \longleftarrow \mathrm{train}(\mathcal{R},\; \{\mathcal{D}_{\mathrm{s}}^\mathrm{valid}:0, \mathcal{D}_{\mathrm{v}}:1\})$\;\label{alg:code:train}
\For{$\mathcal{D}_{\mathrm{t}}\in \mathrm{D}\backslash\{\mathcal{D}_{\mathrm{v}}\}$}{
\tcc{Evaluate accuracy of $r$.}
acc $ \longleftarrow \mathrm{eval}(r,\; \{\mathcal{D}_{\mathrm{s}}^\mathrm{test}:0, \mathcal{D}_{\mathrm{t}}:1\})$\;\label{alg:code:eval}
add acc to $A$\;
}
}
return mean($A$)\;
}
\caption{OD-test}
\label{alg:eval}

\end{algorithm}


As an example, let us walk through the stages of evaluation for \texttt{PbThreshold}, the method that thresholds the maximum probability of a discriminative neural network. We need a trained deep neural network and a threshold for the maximum probability that would reject the outliers. We first train a deep neural network on \D{s} to discriminate the image classes (line~\ref{alg:code:m_train}). The reject class $\mathcal{R}$ returned on line~\ref{alg:code:m_train} will have a single free parameter $\tau$, the threshold to use on the underlying classifier. We pick the optimal threshold $\tau$ in the next step. On line~\ref{alg:code:train}, we pick the best threshold $\tau$ to discriminate between \D{s} and \D{v}.
After finding the best threshold $\tau$ on line~\ref{alg:code:eval}, we evaluate the learned reject function on \D{s} and \D{t}. For unsupervised OOD detection methods the evaluation is a single loop over \D{t} (see supplementary material).
\addtolength{\textfloatsep}{+0.15in}

Note that to score high on OD-test we \textit{do not need to perform density estimation}. However, if we can perform density estimation well, we can score high on OD-test too. If the method $\mathcal{M}$ successfully learns a density function for $\mathcal{D}_s$, we only would have to pick a single threshold to reject OOD samples. Methods that yield a confidence (or uncertainty) estimate can be used similarly. A binary classifier trained with the traditional supervised learning approaches is also not sufficient to score high on this benchmark since we change the second distribution during the test stage. In Section \ref{sec:results} we show how a traditional binary classifier would fall short. The methods that learn conservative boundaries around \D{s} will have a higher chance of success. All the existing approaches can be implemented within this framework.


We equalize the binary classes and only measure accuracy. Furthermore, we require the \textit{methods} to pick the optimal parameters such as the threshold. These choices simplify aggregation, analysis, and comparison of the results. We can meaningfully average over multiple experiments and robustly compare methods in a variety of conditions.
Similar to the supervised learning regime, we can incorporate the prior knowledge on abundance and the risk associated with the OOD samples into the evaluation by modifying the implicit objective on line \ref{alg:code:train} and \ref{alg:code:eval}. We leave the proper choice of application dependent schemes to the practitioner and focus on assessing the discriminative power of the methods in the fixed $50/50$ scenario \textit{without} any prior knowledge. Through this constraint, we ensure that the methods that rely on the prior likelihood cannot perform better than random prediction.

We extend the previous work by evaluating over a broader set of datasets with varying levels of complexity. The variation in complexity allows for a fine-grained evaluation of the techniques. Since OOD detection is closely related to the problem of density estimation, the dimensionality of the input image will be of vital importance in practical assessments. As the input dimensionality increases, we expect the task to become much more difficult. Therefore, to provide a more accurate picture of performance, it is crucial to evaluate the methods on high dimensional data.
Table~\ref{tab:ds_summary} summarizes the datasets that we use.
We also evaluate with uniform, and Gaussian noise for outliers.
We evaluate the following methods:
\begin{itemize}[label={$\bullet$}]
\setlength\itemsep{-0.3em}
    \item \texttt{BinClass}. A traditional binary classifier that is directly trained on \D{s} vs. \D{v}.
    \item \texttt{PbThreshold}. A threshold on the softmax.
    \item \texttt{ScoreSVM}. An SVM~\citep{Cortes1995} classifier on the logits.
    \item \texttt{LogisticSVM}. Similar to \texttt{ScoreSVM}, but the underlying classifier is trained with logistic loss.
    \item \texttt{ODIN}. A threshold on the scaled softmax outputs of the perturbed input.
    \item \texttt{K-NNSVM}. A linear SVM on the sorted Euclidean distance between the input and the k-nearest training samples. Note that a threshold on the average distance is a special case of \texttt{K-NNSVM}.
    \item \texttt{AEThreshold}. A threshold on the autoencoder (AE) reconstruction error of the given input. We train AEs with binary cross-entropy (BCE) and mean squared error (MSE).
    \item \texttt{K-MNNSVM}, \texttt{K-BNNSVM}. A \texttt{K-NNSVM} applied on the learned hidden representations of an autoencoder with MSE or BCE.
    \item \texttt{K-VNNSVM}. Similar to the previous, except we use the learned representation of a variational autoencoder~\citep{Kingma2014}.
    \item \texttt{MC-Dropout}. A threshold on the entropy of average prediction of $7$ evaluations per input.
    \item \texttt{DeepEnsemble}. Similar to \texttt{MC-Dropout}, except we average over the predictions of $5$ networks that are trained independently with adversarial-augmented loss.
    \item \texttt{PixelCNN++}. A threshold on the log-likelihood of each input.
    \item \texttt{OpenMax}. Similar to \texttt{ScoreSVM}, but we use the calibrated output of the OpenMax module that also includes a probability for an unknown class.
\end{itemize}

We use two generic architectures: \texttt{VGG-16}~\citep{Simonyan2014} and \texttt{Resnet-50}~\citep{He2016} and reuse the same base classifier for all the methods to provide a fair and meaningful comparison. We train these architectures with cross-entropy loss (CE), and k-way logistic loss (KWL). CE loss enforces mutual exclusion in the predictions while KWL loss does not. We test these two loss functions to see if the exclusivity assumption of CE hurts the ability to predict OOD samples. CE loss cannot make a \texttt{None} prediction without an explicitly defined \texttt{None} class, but KWL loss can make \texttt{None} predictions through low activations of all the classes.

Note that our formulation of the problem separates the target task from the OOD sample detection. Thus, it is plausible to use, for OOD detection, a different predictive model from the actual predictive model of the target problem if there is an advantage. We tune the hyper-parameters of these methods following the best practices and the published guidelines in the respective articles. The implementation details, a discussion of evaluation cost, and the performance statistics of the above methods are in the supplementary material. The PyTorch implementation with the pre-trained models and all the numerical results is available on \projectPage{}.

\section{Results and Discussion}
\label{sec:results}

\begin{figure}[]
\centering
  \centering
  \includegraphics[width=0.70\textwidth]{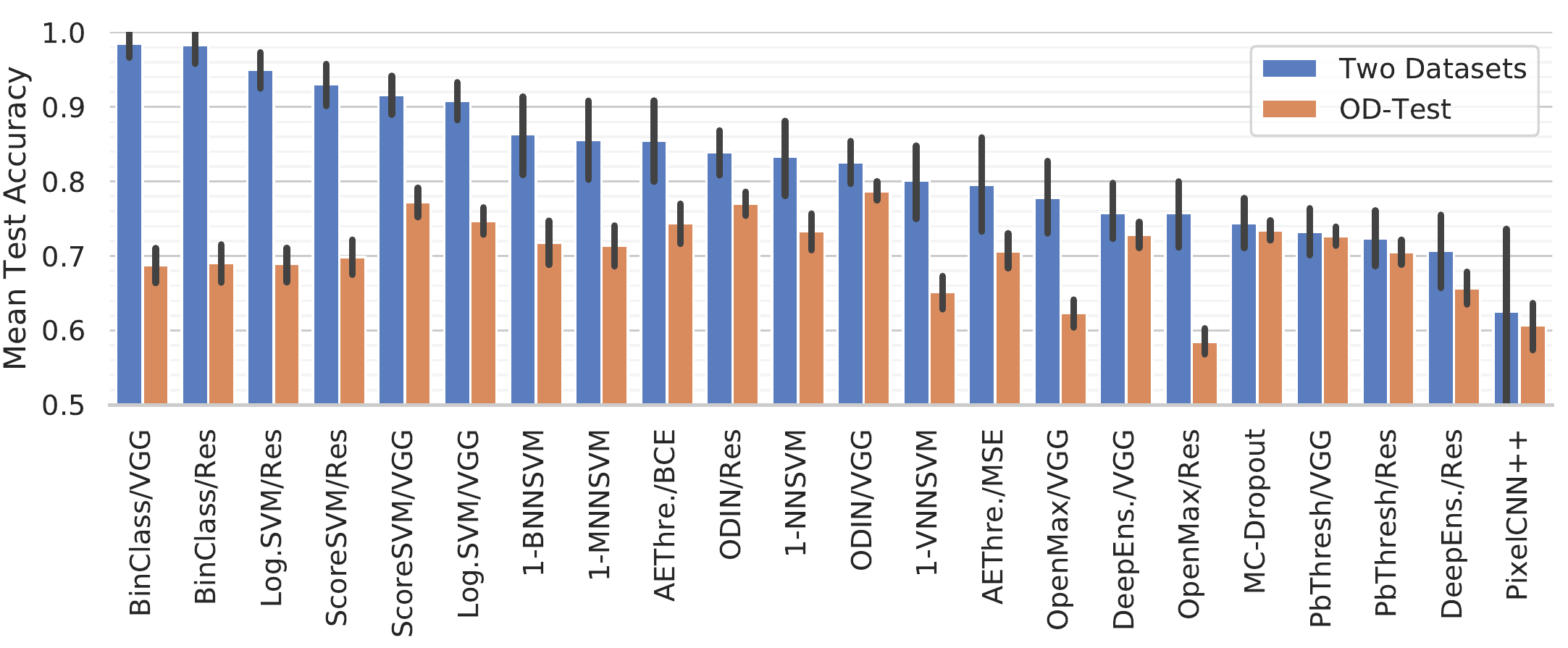}
\caption{Evaluation with two datasets versus OD-test. Evaluating OOD detectors with only two distributions can be misleading for practical applications. The error bars are the 95\% confidence level. The two-dataset evaluations are over all possible pairs of datasets ($n=46$), whereas the OD-test evaluations are over all possible triplets ($n=308$).}
  \label{fig:eval_comp}
\end{figure}

\begin{figure}[]
  \centering
    \centering
    \includegraphics[width=0.70\textwidth]{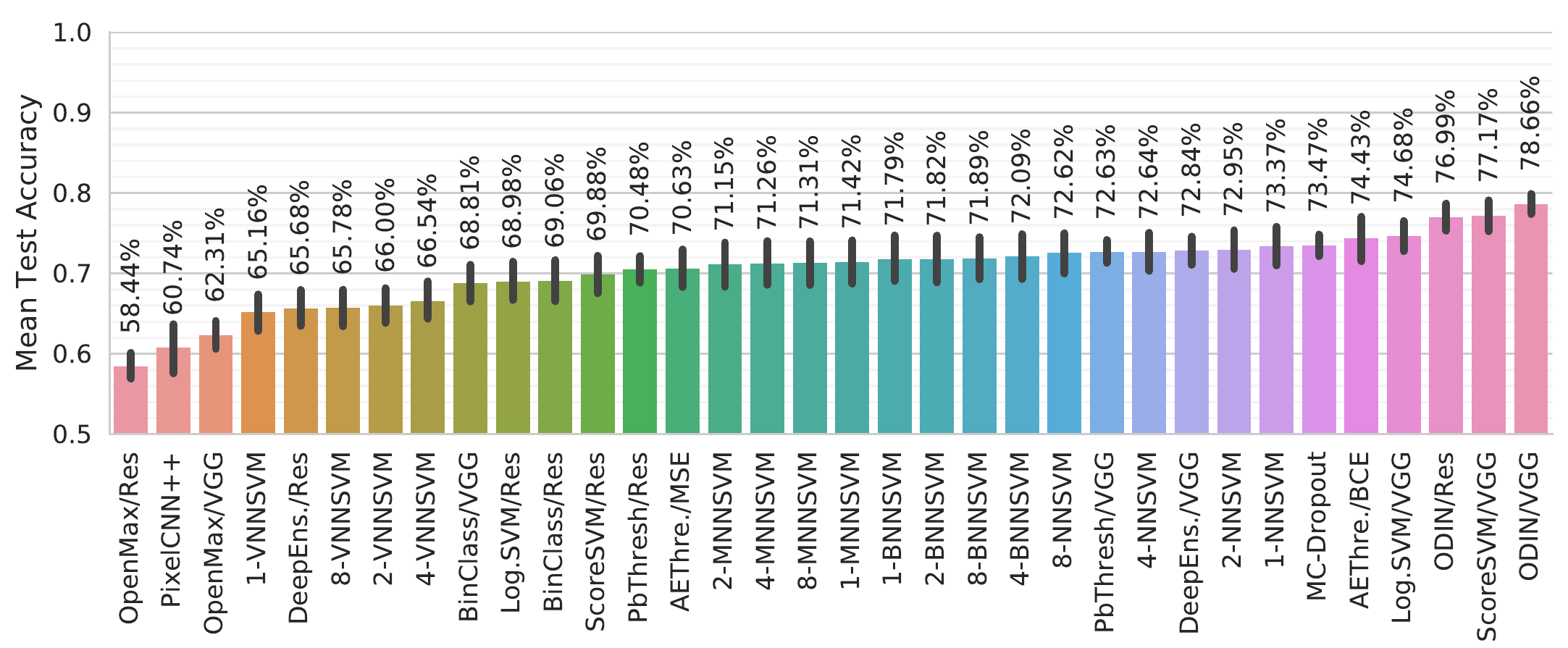}
    \caption{The average test accuracy of the OOD detection methods over $308$ experiments/method with 95\% confidence level. \texttt{/VGG} or \texttt{/Res} indicates the backing network architecture. \texttt{\#-NN./} is the number of nearest neighbours. A random prediction would have an accuracy of $0.5$.}
    \label{fig:mean_overall}
    \vspace{-0.4cm}
\end{figure}

We evaluate the methods in a controlled regime against the datasets in Tab.~\ref{tab:ds_summary}. We run over \num{10000} experiments on all combinations of \D{s}, \D{v}, and \D{t}. First we analyze the aggregated results, then we look at the breakdown of the accuracy per source dataset.
Figure~\ref{fig:eval_comp} and Figure~\ref{fig:mean_overall} show the average accuracy. Each method under OD-test is tested over the same set of $308$ experiments consisting of \textit{all compatible triplet of datasets}. The two-dataset evaluations are averaged over $46$ experiments (all possible pairs). See the project page for the list of experiments.

Figure~\ref{fig:eval_comp} compares the mean test accuracy of methods within OD-test and the two-dataset setting. Methods that perform well in distinguishing two datasets fail when a third dataset is introduced. The gap in relative performance within each evaluation highlights the importance of having a more realistic assessment in practice. The general trend of the evaluation indicates that the methods that have a higher degree of freedom for OOD detection (the space characterized by $\mathcal{R}$) are the most susceptible ones to overestimating accuracy within the traditional evaluation approach. This observation is consistent with the bias-variance tradeoff in learning theory.

\texttt{BinClass}, the direct binary classifier, consistently achieves a near-perfect accuracy on the train and validation of the same datasets (see Fig.~\ref{fig:eval_comp}), however, once tested on a third dataset that it has not seen before, the average accuracy drops to $68\%$ for both \texttt{VGG} and \texttt{Resnet}. This demonstrates why we need to adopt a new evaluation scheme. The classifier overfits to the two distributions (\D{s} and \D{v}), but it cannot distinguish a third distribution~(\D{t}). Because of the diversity of OOD samples, we may always encounter new input that we have not seen before.

\texttt{MC-Dropout} and \texttt{DeepEnsemble}, the two uncertainty techniques, do not seem to provide a strong enough signal to distinguish the two classes compared to the simpler \texttt{ScoreSVM}. Interestingly, \texttt{MC-Dropout} has a higher accuracy than \texttt{DeepEnsemble}. Considering the training cost of \texttt{DeepEnsemble}, \texttt{MC-Dropout} is a more favourable choice.

\texttt{VGG}-backed and \texttt{Resnet}-backed methods significantly differ in accuracy. The gap indicates the sensitivity of the methods to the underlying networks. \texttt{PbThreshold}, \texttt{ScoreSVM}, and \texttt{ODIN} all prefer \texttt{VGG} over \texttt{Resnet} even though \texttt{Resnet} networks outperform the \texttt{VGG} variants in image classification. This means that the image classification accuracy may not be the only relevant factor in performance of these methods. \texttt{ODIN} is less sensitive to the underlying network.
Furthermore, training the networks with KWL loss consistently reduces the accuracy of OOD detection methods on average. \texttt{ScoreSVM/VGG} and \texttt{ScoreSVM/Res} both outperform \texttt{LogisticSVM/VGG}, and \texttt{LogisticSVM/Res} respectively. Similarly, the autoencoders that were trained with BCE loss (\texttt{AEThre./BCE}) outperform the ones trained with MSE loss (\texttt{AEThre./MSE}). Note that we are comparing identical architectures.

Within the nearest-neighbour methods, \texttt{\#-(X)NNSVM}, the number of the nearest neighbours does not significantly impact the accuracy on average. However, performing the nearest-neighbour in the input space directly outperforms nearest-neighbour in the latent representations of autoencoders (\texttt{BNNSVM}, and \texttt{MNNSVM}) and VAE (\texttt{VNNSVM}). Interestingly, \texttt{1-NNSVM} has a higher accuracy than thresholding the probability (\texttt{PbThresh}) and \texttt{DeepEnsemble} on average within OD-test \textit{and} two-dataset evaluation (Fig.~\ref{fig:eval_comp}). For \texttt{\#-NNSVM}, if the reference samples fit the GPU memory, a naive implementation could be faster than a forward pass on the neural networks of large datasets like TinyImagenet.

\texttt{PixelCNN++}, the method that estimates the log-likelihood, has a surprisingly low accuracy on this problem on average. We suspect the auto-regressive nature of the model, specifically when coupled with the IID assumption, may be the reason for its failure. The network approximates the likelihood only in the vicinity of the training data. If we evaluate the model on points that are far from the training data, the estimates are not reliable anymore.

\begin{figure}[]
\centering
  \centering
  \includegraphics[width=0.87\textwidth]{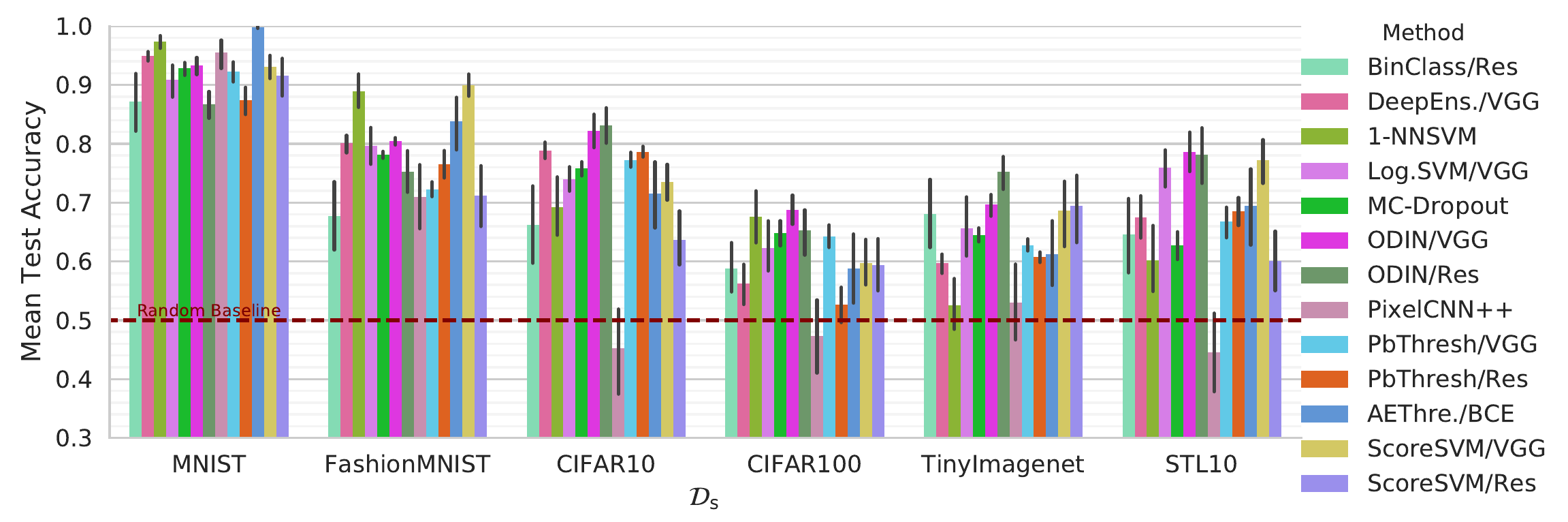}
\caption{The test accuracy over 50 experiments/bar with 95\% confidence level.}
  \label{fig:mean_ds_1}
  \vspace{-0.4cm}
\end{figure}

Figure~\ref{fig:mean_ds_1} shows the average test accuracy across each source dataset \D{s}. For the full figure, see supplementary material. Our quantification of performance shows that all the methods have a much lower accuracy on high-dimensional data than the low-dimensional data.

In low-dimensional datasets, \texttt{K-NNSVM} performs similarly or better than the other methods. In the high-dimensional case, however, the accuracy approaches the random baseline quickly. Interestingly, \texttt{K-NNSVM} performs better on STL10 ($96\times96$) than TinyImagenet ($64\times64$) which might be due to the higher diversity in TinyImagenet compared to STL10. Given the high accuracy of \texttt{K-NNSVM} in $784$ dimensions, it might be feasible to learn an embedding for high-dimensional data, or learn a kernel function, to replace the original image space and enjoy a high accuracy.
Going from CIFAR10 to CIFAR100, the dimensionality and the dataset size remains the same, the only changing factor is the diversity of the data and that seems to make the problem as difficult as the higher dimensional datasets. Except for \texttt{K-NNSVM}, the accuracy of other methods drops significantly in this transition.

\texttt{AEThreshold} has a near perfect accuracy on MNIST, however, the performance drops quickly on complex datasets. \texttt{AEThreshold} also outperforms the density estimation method \texttt{PixelCNN++} on all datasets. We did not explore autoencoder architectures -- more research on better architectures or reconstruction constraints for \texttt{AEThreshold} may potentially have a high pay-off. The top-performing method, \texttt{ODIN}, is influenced by the number of classes in the dataset. Similar to \texttt{PbThreshold}, \texttt{ODIN} depends on the maximum signal in the class predictions, therefore the increased number of classes would directly affect both of the methods. Furthermore, neither of them consistently prefers \texttt{VGG} over \texttt{Resnet} within all datasets.
Overall, \texttt{ODIN} consistently outperforms others in high-dimensional settings, but all the methods have a relatively low average accuracy in the $60\%$-$78\%$ range.

Overall, we can summarize the main observations as follows:
(i)~Outlier detection with two datasets yielded overly optimistic results with \texttt{VGG} and \texttt{Resnet}. (ii)~The \texttt{MC-Dropout} and \texttt{DeepEnsemble} uncertainty methods were not reliable enough for OOD detection.
(iii)~A more accurate image classifier did not lead to a more accurate outlier detector on average.
(iv)~The nearest neighbour methods were competitive in low-dimensional settings, both in computational cost and accuracy. (v) The latent representations of vanilla (variational)-autoencoders were not useful for this task when combined with nearest neighbour methods.
(vi)~The state-of-the-art auto-regressive density estimation method had a surprisingly low accuracy, performing worse than the random prediction baseline in some settings.
(vii)~Although \texttt{ODIN} outperforms other methods in realistic high-dimensional settings, its average accuracy is still below $80\%$. 

To perform supervised OOD sample detection in practice, we have to pick a method and choose a training outlier set \D{v}. Assuming that \D{v} may not represent the full spectrum of anomalies, we should pick methods that do not overfit to \D{v}. OD-test tells us which methods are less likely to overfit to a chosen \D{v} and therefore be more reliable in the face of an unseen OOD sample. Our results show that a two-dataset evaluation scheme can be too optimistic in identifying the best available method. OD-test is a more realistic evaluation of OOD sample detectors.  In practice, the outlier set \D{v} should contain the largest variety of anomalies that we can use, and the method should be the one that is more accurate and less likely to overfit.

\vspace{-0.2cm}
\section{Conclusion}
By detecting OOD samples (the outliers), we can ensure the deep learning pipelines operate as expected.
We assume that collecting a representative set of outliers for \texttt{None} prediction is not always practical.
Therefore, we prefer methods that detect unseen outliers (the unknown unknowns) over methods that only detect previously seen outliers (the known unknowns).
To assess how well methods cope with unknown unknowns we introduced a third dataset in the evaluation of outlier detectors.
Furthermore, we averaged the accuracy across all combinations of outlier datasets to reduce the measurement bias.
We called this evaluation strategy OD-test.
OD-test is a new formulation of the problem that provides a more realistic assessment of the OOD detection methods.
We also presented a new benchmark for OOD sample detection within image classification pipelines based on OD-test.
We showed that the traditional supervised learning approach to OOD detection does not always yield reliable results --
the previous assessments of the OOD detectors are too optimistic to be practical in many scenarios.
We presented a comprehensive evaluation of a diverse set of approaches across a wide variety of datasets for the OOD detection problem.
Furthermore, we showed that \textit{none of the methods is suitable out-of-the-box for high-dimensional images}.
We release the open-source PyTorch project with the pre-trained models to replicate the results of our study.
We invite the community to tackle the outlined challenges in this work.

\vspace{-0.5cm}
\paragraph{Acknowledgments} We would like to thank the reviewers for their helpful comments. We gratefully acknowledge the support of NVIDIA Corporation through the donation of the GPUs used for this research. This work was supported in part by the National Science and Engineering Research Council of Canada (NSERC), and the Canadian Institute for Advanced Research (CIFAR) Learning in Machines and Brains program.

\bibliography{library}
\newpage
\section{Appendix: Formulation and Evaluation}
\subsection{Evaluation of Unsupervised Techniques}
\label{apx:sec:ueval}
\begin{algorithm}
\DontPrintSemicolon
\LinesNumbered
\SetKwInOut{Input}{input}
\SetKwInOut{Output}{}
\SetKwInOut{OptInput}{input}
\Input{$\mathcal{D}_{\mathrm{s}}=(\mathcal{D}_{\mathrm{s}}^\mathrm{train},\mathcal{D}_{\mathrm{s}}^\mathrm{valid},\mathcal{D}_{\mathrm{s}}^\mathrm{test})$ the source dataset.}
\Input{$\mathrm{D}=\{\mathcal{D}_i\}$ outlier \textbf{set}.}
\Input{$\mathcal{M}:\mathcal{D}\rightarrow\mathcal{R}$ the method under evaluation.}
\Begin{
$A \longleftarrow \{\}$\;
\tcc{Generate a rejection hypothesis $r$ using $\mathcal{D}_{\mathrm{s}}^\mathrm{train}$.}
$r \longleftarrow \mathcal{M}(\mathcal{D}_{\mathrm{s}}^\mathrm{train})$\;
\For{$\mathcal{D}_{\mathrm{t}}\in \mathrm{D}$}{
\tcc{Evaluate the accuracy of $r$.}
acc $ \longleftarrow \mathrm{eval}(r,\; \{\mathcal{D}_{\mathrm{s}}^\mathrm{test}:0, \mathcal{D}_{\mathrm{t}}:1\})$\;
add acc to $A$\;
}
return mean($A$)\;
}
\caption{OD-test -- the evaluation procedure for an unsupervised method $\mathcal{M}$.\label{alg:ueval}}
\end{algorithm}

Algorithm~\ref{alg:ueval} outlines the steps of evaluation for an unsupervised method $\mathcal{M}$. Note that in this setting $\mathcal{M}$ returns a single binary classifier $r$. To make the performance of supervised and unsupervised methods comparable, we use the same splits of the datasets to guarantee fairness of evaluation. In this work, we do not evaluate any unsupervised method. This additional information is provided for clarity and completeness.

\subsection{Implementation Details}
\label{apx:sec:impl_details}
For each training procedure, we randomly separate $80\%$ and $20\%$ of the (sub-)data for training and testing respectively. We return the model that has the highest performance on the test (sub-)subset. For classification tasks, we measure the performance by classification accuracy while for other tasks such as \texttt{AEThreshold} we measure the performance through the respective loss value on the test set.

\paragraph{\texttt{VGG}, \texttt{Resnet}} We train two generic classifier architectures \texttt{VGG-16} and \texttt{Resnet-50} on \Ds{s}{train} to perform the corresponding classification task within the datasets. The network architectures slightly differ across datasets to account for the change in the spatial size or the number of classes. We apply our modifications to the reference implementations available in PyTorch's \texttt{torchvision} package. These trained networks are subsequently used in \texttt{PbThreshold}, \texttt{ScoreSVM}, \texttt{ODIN}, \texttt{Log.SVM}, and \texttt{DeepEnsemble}. For \texttt{MC-Dropout} we only use the \texttt{VGG} variant, as the \texttt{Resnet} variants do not have dropouts. Table~\ref{tab:acc_summary} shows the summary of the networks' classification accuracies on the entire \Ds{s}{train} set.

\begin{table*}
\caption{The classification accuracy of the trained networks on \Ds{s}{train} using cross-entropy (CE) and K-way Logistic (KL) loss functions. In both scenarios, the prediction is the maximum activation. Note that because of the difference in training data, this table is not comparable to the state-of-the-art performance on the respective datasets.}
\label{tab:acc_summary}
\small
\centering
\resizebox{\columnwidth}{!}{%

\setlength{\tabcolsep}{4pt}
\begin{tabular}{@{}lccrccr@{}}
\toprule
& \multicolumn{3}{c}{\texttt{VGG}} &
  \multicolumn{3}{c}{\texttt{Resnet}} \\
\cmidrule(r{4pt}){2-4} \cmidrule(l){5-7}
& CE-Accuracy & KL-Accuracy & Size & CE-Accuracy & KL-Accuracy & Size \\
\midrule
MNIST~\citep{LeCun1998}              & 99.89\% & 99.91\% & 19\,MB & 99.89\% & 99.91\% & 70\,MB\\
FashionMNIST~\citep{Xiao2017}        & 98.82\% & 98.36\% & 19\,MB & 98.75\% & 98.73\% & 70\,MB\\
CIFAR10~\citep{Krizhevsky2009}       & 97.63\% & 97.34\% & 159.8\,MB & 97.75\% & 97.51\% & 94.3\,MB\\
CIFAR100~\citep{Krizhevsky2009}      & 91.40\% & 91.87\% & 161.3\,MB & 92.05\% & 91.82\% & 95.1\,MB\\
TinyImagenet\footnote{https://tiny-imagenet.herokuapp.com/} & 69.71\% & 72.23\% & 162.9\,MB & 89.59\% & 65.95\% & 95.9\,MB\\ 
STL10~\citep{Coates2011} & 93.62\% & 95.18\% & 201.7\,MB & 92.32\% & 93.34\% & 94.3\,MB\\
\midrule
Mean & 91.84\% & 92.48\%& & 95.05\%& 91.21\%& \\
\bottomrule
\end{tabular}
}
\end{table*}

\paragraph{Data Augmentation} We only allow mirroring of the images for data augmentation. We apply data augmentation on all the datasets except \texttt{MNIST} for which mirror augmentation does not make sense. We explicitly instantiate mirrored samples, as opposed to implicit on-air augmentation, to ensure methods such as \texttt{K-NNSVM} are not at disadvantage. We do not apply any other data augmentation. We initially experimented with no data augmentation. Without any augmentation, the performance of all the methods reduces by $3-4\%$, but the relative ranking stays the same.

\paragraph{\texttt{BinClass}} A binary classifier that is directly trained on \D{s} and \D{v}. We use the reference \texttt{Resnet} or \texttt{VGG} architectures with an additional linear layer to transform the output of the network to a one-dimensional activation. We train the network with a binary cross-entropy loss on (\Ds{s}{train}+\Ds{s}{valid}:0, \D{v}:1) to ensure the method has access to the same data as the other methods. The networks typically achieve near-perfect accuracy after only a few epochs.

\paragraph{\texttt{PbThreshold}~\citep{Hendrycks2017}} One threshold parameter on top of the maximum of softmax output. The cost of the evaluation is a single forward pass on the network. We reuse the trained reference \texttt{VGG} or \texttt{Resnet} architectures for this.

\paragraph{\texttt{ScoreSVM}} A natural generalization of \texttt{PbThreshold} is to train an SVM~\citep{Cortes1995} classifier on the pre-softmax activations. The cost of evaluation is a single forward pass on the network with the additional SVM layer. We reuse the trained reference \texttt{VGG} or \texttt{Resnet} architectures for this. We set the weight-decay regularization to $\frac{1}{m}$, where $m$ is the size of the training set.

\paragraph{\texttt{ODIN}~\citep{Liang2018}} A threshold on the softmax outputs of the perturbed input. The cost of the evaluation is two forward passes and one backward pass. We do a grid search over the $\epsilon$, the perturbation step size, and $\gamma$, the temperature of the softmax operation. The range for grid search is the same as the suggested range in \cite{Liang2018}. We reuse the trained reference \texttt{VGG} or \texttt{Resnet} architectures for this.

\paragraph{\texttt{K-NNSVM}} A linear SVM on the sorted Euclidean distance between the input and the k-nearest training samples. Note that a threshold on the average distance is a special case of \texttt{K-NNSVM}. The cost of the evaluation is finding the k-nearest neighbours in the training data. We use the \Ds{s}{train} as the reference set, and tune the parameters with (\Ds{s}{valid}, \D{v}).

\paragraph{\texttt{K-MNNSVM}, \texttt{K-BNNSVM}, \texttt{K-VNNSVM}} The same as \texttt{K-NNSVM}, except we use the low dimensional representations of an autoencoder trained with MSE, BCE, or the VAE.

\paragraph{\texttt{AEThreshold}} A threshold on the autoencoder reconstruction error of the given input. The evaluation cost is a single forward pass on the autoencoder. We train the autoencoder on \Ds{s}{train} and train the threshold parameters with (\Ds{s}{valid},\D{v}). We use the binary cross-entropy loss with continuous targets\footnote{See \url{http://pytorch.org/docs/0.3.1/nn.html\#bcewithlogitsloss}.} or mean squared error to train and measure the reconstruction error of a given input. The bottleneck dimensionality varies between $32$ and $1024$. Our decision rule given a reconstruction error $e_x$ for an input $x$ is $r(x) = (e_x-\mu)^2 > \tau$, where $\tau$ is the threshold and $\mu$ is the center around which we are thresholding with $\tau$. If we set $\mu=0$, this decision function reduces to a basic threshold operator. We found that this simple decision rule improves the final accuracy of the model. The reconstruction errors of the in-distribution samples tend to stay more or less similar, whereas the reconstruction error for OOD samples could either be too low or too high. This decision rule is meant to utilize this observation. The network architectures are procedurally generated. See \projectPage{/blob/master/models/autoencoders.py} for the models.

\paragraph{\texttt{MC-Dropout}} A threshold on the entropy of average predictions of $7$ evaluations per input. The dropout probability is $p=0.5$. This approach follows the work of \cite{Lakshminarayanan2017} and Kendall and Gal~\citep{Kendall2017}. We did not evaluate this approach on \texttt{Resnet} because the original structure does not have a dropout; therefore, it is not trivial to identify where the dropouts should be located without sabotaging the performance of \texttt{Resnet}. We reuse the trained reference \texttt{VGG} architecture for this.

\paragraph{\texttt{DeepEnsemble}} Similar to \texttt{MC-Dropout}, except we average over the predictions of $5$ networks that are trained independently with the adversarial strategy of \cite{Lakshminarayanan2017}. In this approach, we augment the original loss function with a similar loss function on the adversarially-generated examples of the same batch. The adversarially-generated samples are generated through the fast gradient-sign method (FGSM)~\citep{Goodfellow2015}.

\begin{figure}
\ffigbox[\textwidth]
{
    \begin{floatrow}
    \ffigbox[\linewidth]{%
        \captionof{subfigure}{MNIST}
    }{
        \includegraphics[width=0.4\textwidth]{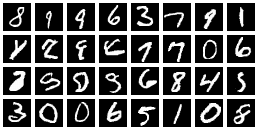}
    }
    \ffigbox[\linewidth]{%
        \captionof{subfigure}{FashionMNIST}
    }{
        \includegraphics[width=0.4\textwidth]{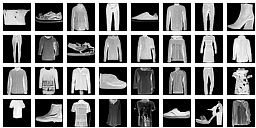}
    }
    \end{floatrow}
    
    \begin{floatrow}
    \ffigbox[\linewidth]{%
        \captionof{subfigure}{CIFAR10}
    }{
        \includegraphics[width=0.4\textwidth]{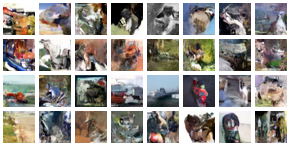}
    }
    \ffigbox[\linewidth]{%
        \captionof{subfigure}{CIFAR100}
    }{
        \includegraphics[width=0.4\textwidth]{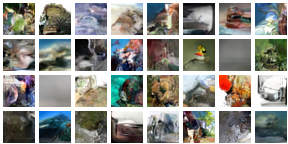}
    }
    \end{floatrow}
    \begin{floatrow}
    \ffigbox[\linewidth]{%
        \captionof{subfigure}{TinyImagenet (downsampled)}
    }{
        \includegraphics[width=0.4\textwidth]{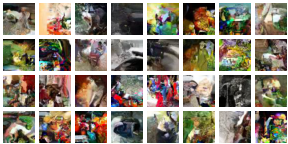}
    }
    \ffigbox[\linewidth]{%
        \captionof{subfigure}{STL10 (downsampled)}
    }{
        \includegraphics[width=0.4\textwidth]{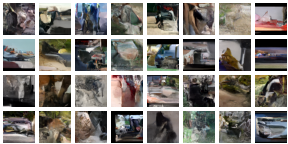}
    }
    \end{floatrow}
}{
\caption{Samples of the learned \texttt{PixelCNN++} model.}\label{apx:fig:pcnn_samples}
}
\end{figure}

\paragraph{\texttt{PixelCNN++}} We use the implementation from \url{https://github.com/pclucas14/pixel-cnn-pp}. We train the models using \D{s} until plateau on the test (sub-)subset, then learn a threshold parameter with \D{v}. Our models achieve a 0.89 BPD for MNIST, 2.65 BPD for FashionMNIST, 2.98 BPD for CIFAR10, 3.01 BPD for CIFAR100, 2.70 BPD for TinyImagenet, and 3.59 BPD for STL10 on the test (sub-)subset. Because of the auto-regressive nature, these models are prohibitively expensive to train. The \texttt{PixelCNN++} authors note that they have used 8 Titan X GPUs for five days to achieve state-of-the-art performance for CIFAR10\footnote{\url{https://github.com/openai/pixel-cnn}} (2.92 BPD). For TinyImagenet, and STL10 we process a downsampled version to 32-pixel width to be able to train and evaluate the models. Our experiments with \texttt{AEThreshold} indicate that the downsampled versions of TinyImagenet, and STL10 are easier problems. However, even with this simplification, the \texttt{PixelCNN++} does not perform up to expectations. Figure~\ref{apx:fig:pcnn_samples} shows some of the generated samples.

\paragraph{\texttt{OpenMax}} This method is a replacement for the softmax layer \textit{after} the training has finished. It fits a Weibull distribution on the distances of logits from the representatives of each class to reweight the logits and provide probabilities for encountering an unknown class. The output of the OpenMax is similar to softmax, except with the addition of the probability for an unknown class. We learn the MAV vectors and the Weibull distribution on the \D{s}. We use the \D{v} to learn the reject function on the calibrated probability outputs.

\vspace{1cm}
\noindent{}You can access all the results on \projectPage{} where you will find the full list of evaluations for OD-test ($n=34\times308=\num{10472}$) and the two-dataset evaluation scheme ($n=22\times46=\num{1012}$).

\section{Appendix: More Results}
\label{apx:sec:results}
\begin{sidewaysfigure*}
\centering
  \centering
  \includegraphics[width=\textwidth]{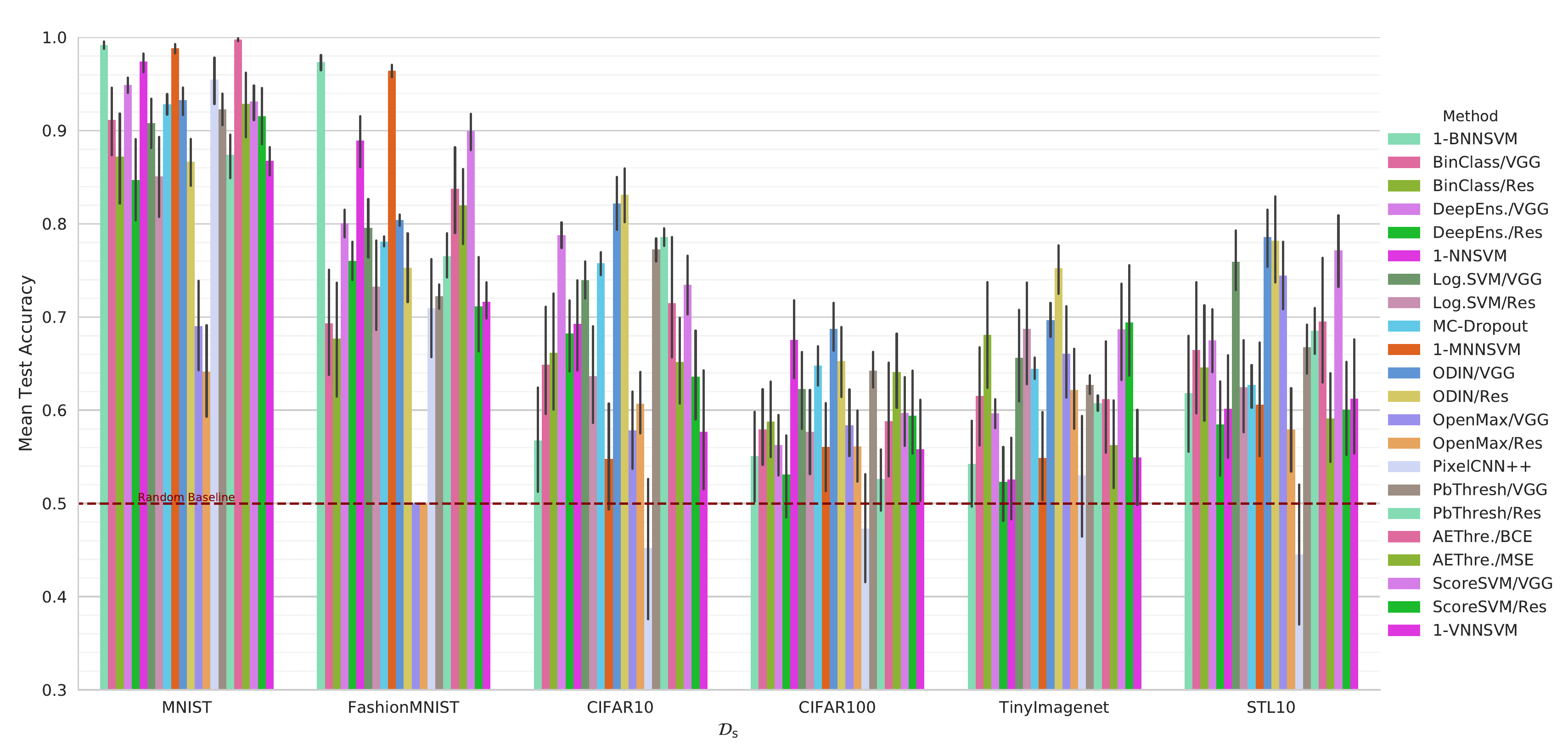}
\captionof{figure}{The average test accuracy over 50 experiments per bar. The error bars indicate the 95\% confidence level. The figure is best viewed in color.
  }
  \label{fig:mean_ds_all}
\end{sidewaysfigure*}

Figure~\ref{fig:mean_ds_all} shows the average performance of all the methods per source dataset \D{s}.

\end{document}